\documentclass[11pt, a4paper, copyright, gdm]{google}

\usepackage[authoryear, sort&compress, round]{natbib}
\usepackage{nicefrac}

\usepackage{listings}
\usepackage{upquote}
\definecolor{promptframe}{HTML}{B4B2AB}
\lstdefinestyle{prompt}{
  basicstyle=\ttfamily\scriptsize,
  breaklines=true,
  breakautoindent=false,
  columns=fullflexible,
  keepspaces=true,
  frame=single,
  framesep=5pt,
  rulecolor=\color{promptframe},
  xleftmargin=2pt,
  xrightmargin=2pt,
  aboveskip=6pt,
  belowskip=6pt,
}
\usepackage[section,below]{placeins}
\newcount\Comments
\newif\ifshowcomments
\showcommentsfalse
\definecolor{darkred}{rgb}{0.9,0,0}
\definecolor{darkgreen}{rgb}{0,0.5,0}
\definecolor{darkblue}{rgb}{0,0,0.7}
\definecolor{purple}{rgb}{.6, 0,.6}
\definecolor{orange}{rgb}{1.0,0.64,0}
\newcommand{\kibitz}[2]{\ifshowcomments\ifnum\Comments=1\textcolor{#1}{#2}\fi\fi}

\ifXeTeX
  \edef\ttclrm{\rmdefault}
  \edef\ttclsf{\sfdefault}
  \edef\ttcltt{\ttdefault}
  \usepackage[no-math]{fontspec}
  \usepackage{xeCJK}
  \setCJKsansfont{FandolHei-Regular.otf}
  \setCJKmonofont{FandolFang-Regular.otf}
  
  \renewcommand{\rmdefault}{\ttclrm}
  \renewcommand{\sfdefault}{\ttclsf}
  \renewcommand{\ttdefault}{\ttcltt}
  \normalfont
  \DeclareMicrotypeSet*[protrusion]{ttcltext}{
    encoding = {OT1,T1,LY1,OT4,QX,T5,TU},
    family   = {rm*,sf*,tt*},
  }
\fi

\correspondingauthor{xiaoyin.chen@mila.quebec, hopeyao@google.com}

\uselogo{}

\title{Are Self-Evolving Agents Continual Learners?}

\author[1,2,*]{Xiaoyin Chen}
\author[3]{Yizheng Huang}
\author[3]{Yinglun Xu}
\author[3]{Ziyun Zhao}
\author[3]{Emmett McQuinn}
\author[3]{Julie Liu}
\author[3]{Hope Yao}

\affil[1]{Mila -- Quebec AI Institute}
\affil[2]{Université de Montréal}
\affil[3]{\thepa{}}
\affil[*]{Work done as a student researcher at Google DeepMind}

\begin{abstract}
Large language model agents are often deployed in environments where
new knowledge, tools, and workflows keep arriving. Such an agent is
therefore expected to adapt continually, learning new tasks as they
arrive without forgetting earlier ones. This resembles the continual
learning problem, but at a pace that conventional gradient updates
cannot sustain. Nonparametric self-evolving agents are a natural
alternative: they adapt by rewriting their own harness, such as prompts,
memory, and skills, from task feedback while the model weights stay
frozen. We study whether such agents are effective continual learners
across four drifting task streams that shift in topic, workflow, or
domain (built from BIRD, BFCL, and ScienceWorld). Evaluating three
representative methods spanning experience retrieval and
self-modification (ReasoningBank, GEPA, and Meta-Harness), we find that
although self-evolution improves aggregate performance on most streams,
higher plasticity consistently comes at the cost of stability.
Specifically, self-modifying methods apply
global edits to a shared prompt or workspace, interfering with unseen
tasks and overwriting earlier knowledge as the stream shifts. Conversely,
similarity-based memory retrieval localizes updates across distant
domains, but still retrieves foreign memories on unseen tasks and
suffers from retrieval collision when similar queries require
conflicting workflows. These findings show that current self-evolving
agents remain incomplete continual learners.
\end{abstract}

\begin{document}

\maketitle

\section{Introduction}
\label{sec:intro}

Large language models (LLMs) are increasingly used as agents that act in
interactive environments: they operate software
tools and systems~\citep{yao2023react, patil2025bfcl, zhanginfragym}, navigate embodied
simulators~\citep{shridhar2020alfworld0, wang2022scienceworld0}, and
query databases or write code on behalf of
users~\citep{li2024can, wang2025codecontests00}. The skills such an agent requires
are not fixed at training time. New
tasks, tools, and workflows continue to emerge after the weights are
frozen, and an agent that serves users over an extended period must
learn each new task as it arrives while preserving what it has already
learned. This form of learning resembles the classical continual
learning problem, which studies how a model can be trained on new tasks
over time without forgetting earlier ones~\citep{parisi2019continual, delange2021continual,
vandeven2019three}. Continual learning has recently emerged as a central
challenge for LLMs~\citep{ke2023continual, jang2022towards,
luo2023empirical, shenfeld2026self0distillation, shi2024continual,
wu2024continual}. Conventionally, it has been realized through gradient
updates. An LLM agent, however, is expected to adapt at the timescale of use: every
interaction with the environment may yield knowledge that is useful in a
later session, and that knowledge should be available as soon as the
next session begins. Performing gradient updates at this pace is challenging:
retraining an LLM after every session is prohibitively expensive, and
fine-tuning on a small number of recent interactions is known to erode
general capabilities~\citep{luo2023empirical}.

The in-context learning ability of
LLMs~\citep{brown2020language, dong2022survey} suggests an alternative
route to adaptation. Rather than the weights, an agent can update its
\emph{harness}, everything wrapped around the model: the system prompt,
the accumulated memory, the skill libraries, and the control code that
orchestrates them. \emph{Nonparametric} self-evolving
agents adopt precisely this strategy, refining the harness from task
outcomes and environment feedback while the weights remain frozen. Two
paradigms have emerged. \emph{Self-modification} rewrites the artifacts
on which the agent runs, such as its system prompt or the code of the
harness itself~\citep{zhou2022large, pryzant2023automatic,
agrawal2025gepa, lee2026meta0harness0}; \emph{experience augmentation}
appends past reflections, successful trajectories, or tool libraries to
an external store and retrieves them during
execution~\citep{wang2023voyager, zhao2024expel, ouyang2025reasoningbank0}.
In either paradigm the harness constitutes a state carried across tasks;
updating it is inexpensive, takes effect immediately, and leaves the
model weights untouched. Self-evolving agents are therefore natural
candidates for continual learning at inference time.

Both paradigms demonstrate substantial gains on stationary, single-domain
benchmarks, yet whether they support continual learning has not been examined.
Iterative feedback may drive an agent toward narrow, task-specific
shortcuts, such as idiosyncratic formatting templates or
environ\-ment-specific action sequences, rather than transferable
problem-solving strategies. Real-world scenarios, moreover, extend beyond
a single benchmark: the agent faces a stream of heterogeneous tasks
served by one shared harness, and every update made for one task applies
to all tasks that follow. Whether self-evolution under these conditions
accumulates durable gains or compounding interference is unknown.
In short: \emph{Are current self-evolving LLM agents effective continual
learners?}

We address this question by evaluating the continual learning dynamics
of self-evolving agents on non-stationary task streams. After
formalizing the \emph{evolver}, the function that updates the harness
from collected experience, together with task streams over multiple task
groups (\S\ref{sec:background}), we construct four \emph{drifting}
stream scenarios that train one task group at a time in contiguous
blocks (\S\ref{sec:method}, \S\ref{sec:setup}). Built from
BIRD~\citep{li2024can}, BFCL~\citep{patil2025bfcl}, and
ScienceWorld~\citep{wang2022scienceworld0}, these streams isolate shifts
in subject matter (\emph{topic drift}), interaction workflow
(\emph{workflow drift}), environment and action space (\emph{domain
drift}), and their sequential combination (\emph{compound drift}). We
evaluate three representative self-evolving methods that span both
paradigms and progressively larger harness surfaces:
ReasoningBank~\citep{ouyang2025reasoningbank0} retrieves accumulated
memory items, GEPA~\citep{agrawal2025gepa} rewrites a strategy section
in the system prompt, and Meta-Harness~\citep{lee2026meta0harness0}
edits the system prompt, memory and skill files, and controller code. To
characterize these dynamics, we measure terminal and average advantage
over the seed harness (\S\ref{sec:metrics}) and decompose stream
performance into within-block \emph{plasticity} ($P$) and cross-block
\emph{stability} ($S$), combining pre-block interference ($\bar{I}$) on
unseen groups and post-block forgetting ($\bar{F}$) on earlier groups.

Three main findings emerge:
\begin{itemize}
    \item \textbf{Current methods trade stability for plasticity
    (\S\ref{sec:main-results}).} While self-evolution improves overall
    stream performance in most settings, no method achieves both high
    plasticity and high stability: expanding the editable surface from
    retrieved memories (ReasoningBank) to the system prompt (GEPA) and
    full workspace (Meta-Harness) increases plasticity or edit freedom at
    the cost of cross-block stability.
    \item \textbf{Self-modification causes cross-task interference and
    overwriting (\S\ref{sec:transfer}).} Because GEPA and Meta-Harness
    apply a single shared prompt or workspace to every incoming task,
    task-specific rules specialized to one block interfere with unseen
    tasks before their block begins and are compressed or overwritten by
    later blocks.
    \item \textbf{Similarity-based retrieval cannot prevent interference
    or collision (\S\ref{sec:forgetting}).} Retrieving experiences by
    query similarity (ReasoningBank) avoids post-block forgetting across
    distant domains, but still injects foreign memories into unseen task
    groups and causes retrieval collision when distinct workflows share
    the same domain and embedding space.
\end{itemize}

Our results show that current self-evolving agents remain incomplete
continual learners, calling for mechanisms that isolate task-specific
updates in the harness and condition retrieval beyond surface query
similarity.

\section{Background}
\label{sec:background}

\subsection{Continual Learning over Task Streams}
\label{sec:cl-background}

We define a \emph{task family} $\mathcal{F}$ as a tuple
$\mathcal{F}=(\mathcal{X}, \mathcal{D}, \mathcal{Y}, R)$, where
$\mathcal{X}$ is an input space (e.g., a query), $\mathcal{D}$ is a
distribution over $\mathcal{X}$ from which inputs are sampled,
$\mathcal{Y}$ is an output space (e.g., a response trajectory), and
$R: \mathcal{X} \times \mathcal{Y} \to \mathbb{R}$ is a reward function
that measures the quality of a response trajectory given a query. A
stochastic policy $\pi: \mathcal{X} \to \Delta(\mathcal{Y})$ defined on
the task family $\mathcal{F}$ takes a query $x \in \mathcal{X}$ as input
and outputs a response trajectory $y \sim \pi(\cdot \mid x)$, and the
corresponding reward signal is $r = R(x, y)$. In the agentic setting
studied here, $y$ is an \emph{episode}, the full interaction trajectory
generated in the task environment, and the reward is binary task
success, $R(x, y) \in \{0, 1\}$.

A continual learning process is defined on $K$ task families
$\mathcal{F}_k = (\mathcal{X}_k, \mathcal{D}_k, \mathcal{Y}_k, R_k)$,
$k \in [K]$, and a weight schedule $w = (w_t)_{t=0}^{T-1}$ with
$w_t = (w_{t,1}, \dots, w_{t,K})$ and $\sum_{k=1}^{K} w_{t,k} = 1$. A
\emph{task stream} $S_w = (x_t)_{t=0}^{T-1}$ is a sequence of inputs
drawn from the mixture over the families that $w$ specifies at each
position,
\begin{equation}
  x_t \sim P_t \;:=\; \sum_{k=1}^{K} w_{t,k}\, \mathcal{D}_k,
  \qquad t \in \{0, \dots, T-1\}.
  \label{eq:stream}
\end{equation}
At each timestep $t = 0, \dots, T-1$, the continual learner receives the
input $x_t$ from the stream, generates an output
$y_t \sim \pi^{(t)}(\cdot \mid x_t)$ with its current policy, observes a
reward $r_t$, and then outputs the policy $\pi^{(t+1)}$ for the next
timestep. The average performance of a policy $\pi$ over the $K$ families
is
\begin{equation}
  J(\pi) \;=\; \frac{1}{K} \sum_{k=1}^{K} J_k(\pi),
  \qquad
  J_k(\pi) \;=\; \mathbb{E}_{x \sim \mathcal{D}_k,\; y \sim \pi(\cdot \mid x)}
  \big[R_k(x, y)\big].
  \label{eq:score}
\end{equation}
The continual learner seeks to maximize the expected performance
averaged over the whole stream,
\begin{equation}
  \bar{J}_w \;:=\; \mathbb{E}_{S_w}\Big[\frac{1}{T}
  \sum_{t=1}^{T} J\big(\pi^{(t)}\big)\Big],
  \label{eq:cl-objective}
\end{equation}
where the expectation is over the draw of the stream $S_w$.

This objective formalizes the tension between plasticity and forgetting.
At every position, $J$ evaluates the policy on all $K$ families, not
only the one supplying the current input. Each update should therefore
improve the policy on the families it is currently learning from without
degrading it on families seen earlier: a degraded family lowers the
average at every remaining position until it is relearned.

\subsection{Nonparametric Self-Evolving Agents}
\label{sec:se-form}
We study how a self-evolving agent can serve as the continual learner
above, with its self-evolution step generating the update from
$\pi^{(t)}$ to $\pi^{(t+1)}$. The agent implements this update through
an \emph{evolver} $f$. Let
$\mathcal{H}_t := \{(x_\tau, y_\tau, r_\tau)\}_{\tau=0}^{t}$ denote the
experience collected up to timestep $t$. After each episode, the evolver
maps the current policy and the accumulated experience to the next
policy,
\begin{equation}
  \pi^{(t+1)} = f\big(\pi^{(t)},\; \mathcal{H}_t\big).
  \label{eq:evolve}
\end{equation}
Note that the evolver may use only a subset of the history to generate
an update, which is usually the case in current self-evolving methods.
Existing self-evolving methods are formulated on a single task family,
which corresponds to the special case $K = 1$ of the stream above; we
apply the same update on a stream, with $x_t$ drawn from $P_t$.

In the LLM-agent setting, a policy is an agent determined by an LLM with
parameters $\theta$ and a \emph{harness} $c$: the system prompt,
accumulated memory, skill libraries, and the control code that
orchestrates them. Given an input $x$, rolling out the model in the task
environment yields an episode $y \sim \pi_{\theta}(\cdot \mid x, c)$; we
write this episode-level policy $\pi_{\theta, c}$. An evolver may update
either component. In this work, we study the nonparametric case, in which
$\theta$ is frozen and only the harness is updated, such as in
GEPA~\citep{agrawal2025gepa}, Meta-Harness~\citep{lee2026meta0harness0},
and ReasoningBank~\citep{ouyang2025reasoningbank0}, so that
Eq.~\ref{eq:evolve} reduces to
\begin{equation}
  c_{t+1} = f\big(c_t,\; \mathcal{H}_t\big).
  \label{eq:harness-update}
\end{equation}

On a stream, the evolver traverses the inputs in arrival order and
updates the harness as in Eq.~\ref{eq:harness-update}, yielding
$c_0, c_1, \dots, c_T$ with $\pi^{(t)} = \pi_{\theta, c_t}$ throughout;
the seed harness $c_0$ defines the initial policy $\pi^{(0)}$. We define
the performance of $f$ under a schedule $w$ as the objective of
Eq.~\ref{eq:cl-objective} evaluated on the policy sequence that $f$
generates,
\begin{equation}
  \bar{J}_w(f) \;:=\; \mathbb{E}_{S_w,f}\Big[\frac{1}{T}
  \sum_{t=1}^{T} J\big(\pi^{(t)}\big)\Big].
  \label{eq:stream-score}
\end{equation}

\section{Methodology}
\label{sec:method}

In this work, we study two factors of the problem, the evolver $f$ and
the weight schedule $w$ of the task stream, and we evaluate an evolver
by how much it improves the objective $\bar{J}_w(f)$ of
Eq.~\ref{eq:stream-score} over the seed harness.

\subsection{Task Stream Configurations}
\label{sec:streams}
We instantiate the task stream as a \emph{drifting} stream, in which
$w_t$ is one-hot and piecewise-constant: the stream is a sequence of
contiguous blocks, each drawing all of its tasks from a single task
group, and the active group changes only at block boundaries. Within a
block, every update is driven by the active group, while earlier groups
receive no further tasks and hence no further feedback, and later groups
have not yet been encountered. In continual deployment, transitions
between blocks may alter the subject matter (topic drift), the
interaction workflow required for a correct response (workflow drift),
the environment and action space itself (domain drift), or a combination
of all three (compound drift); we detail the concrete benchmark streams
that instantiate these four drift scenarios in \S\ref{sec:setup}.

In every stream the method maintains one shared harness across the
entire horizon. The stream never reveals the group label of a task or
the block boundaries, and the agent is never told that it operates in a
drifting stream or that it should balance performance across groups. The
agent must infer the structure of incoming tasks on its own and encode
any distinction it finds in the harness, mirroring task-free continual
learning scenarios~\citep{aljundi2019taskfree, delange2021continual,
buzzega2020dark}.

\subsection{Metrics}
\label{sec:metrics}
We evaluate self-evolving agents by how much evolution improves the
objective $\bar{J}_w(f)$ of Eq.~\ref{eq:stream-score} over the seed
harness. Because evaluating the harness after every update is costly, we
compute accuracy at $M$ checkpoints $0 < t_1 < \dots < t_M = T$ along
with the seed position $t_0 = 0$. For a stream position $t$ and a family
$k$, let $A[t, k]$ denote the accuracy (in \%) of the frozen harness
$c_t$ on a disjoint hold-out set $\mathcal{D}_k^{\text{eval}}$ of family
$k$:
\begin{equation}
  A[t, k] \;=\; \frac{1}{|\mathcal{D}_k^{\text{eval}}|}
  \sum_{x \in \mathcal{D}_k^{\text{eval}}}
  \mathbb{E}_{y \sim \pi_{\theta, c_t}(\cdot \mid x)} \big[R_k(x, y)\big].
  \label{eq:acc-matrix}
\end{equation}
The seed harness $c_0$ gives the per-family baseline $B_k := A[0, k]$
before any evolution. Offsetting $A[t, k]$ against $B_k$ isolates what
evolution contributes, reported in percentage points (pp.). We summarize
overall stream performance with two quantities: the \emph{terminal
advantage} ($\text{Term.}$), which averages $A[T, k] - B_k$ over all $K$
families at the end of the stream, and the \emph{average advantage}
($\bar{\Delta}$), which averages $A[t_m, k] - B_k$ over all $K$ families
and all $M$ post-seed checkpoints:
\begin{equation}
  \bar{\Delta} \;=\; \frac{1}{M K} \sum_{m=1}^{M} \sum_{k=1}^{K}
  \big(A[t_m, k] - B_k\big),
  \qquad
  \text{Term.} \;=\; \frac{1}{K} \sum_{k=1}^{K}
  \big(A[T, k] - B_k\big).
  \label{eq:avg-advantage}
\end{equation}

Overall advantage measures the size of the improvement but not its
composition: the same value can come from balanced gains on every family
or from gains on the active family paid for by losses on the others. In
a drifting stream where the block of family $k$ occupies positions
$\sigma_k < t \le \tau_k$, we decompose performance into
\emph{plasticity} and \emph{stability}:
\begin{itemize}
  \item \emph{Plasticity} $P$ measures how much each family learns
  during its own training block~\citep{riemer2018learning,
  kemker2018measuring}, defined as the advantage over the seed at
  the end $\tau_k$ of family $k$'s block, $P_k = A[\tau_k, k] - B_k$,
  averaged over all $K$ families:
  \begin{equation}
    P \;=\; \frac{1}{K} \sum_{k=1}^{K} \big(A[\tau_k, k] - B_k\big).
    \label{eq:plasticity}
  \end{equation}
  \item \emph{Stability} $S$ measures the degradation a family incurs
  while the stream trains the other
  families~\citep{lopezpaz2017gradient, kemker2018measuring}. It combines \emph{pre-block
  interference} $I_k$, the mean drop below the seed at checkpoints
  before family $k$'s block begins ($\sigma_k$), and \emph{post-block
  forgetting} $F_k$, the drop from the end of family $k$'s block
  ($\tau_k$) to the end of the stream ($T$):
  \begin{equation}
    I_k \;=\; \frac{1}{M_k} \sum_{0 < t_m \le \sigma_k}
    \max\big(0,\, B_k - A[t_m, k]\big),
    \qquad
    F_k \;=\; \max\big(0,\, A[\tau_k, k] - A[T, k]\big),
    \label{eq:interf-forget}
  \end{equation}
  where $M_k$ is the number of checkpoints with $0 < t_m \le \sigma_k$
  (with $I_k = 0$ when $M_k = 0$, and $F_k = 0$ when $\tau_k = T$).
  Averaging $I_k$ and $F_k$ over the $K$ families gives stream-level
  interference $\bar{I}$ and forgetting $\bar{F}$, and
  \begin{equation}
    S \;=\; -\frac{1}{K} \sum_{k=1}^{K} \big(I_k + F_k\big)
    \;=\; -\big(\bar{I} + \bar{F}\big).
    \label{eq:stability}
  \end{equation}
\end{itemize}

\section{Experiments}
\label{sec:experiments}

\subsection{Experimental Setup}
\label{sec:setup}

\paragraph{Datasets and streams.}
We construct task streams from three agentic benchmarks:
BIRD~\citep{li2024can}, text-to-SQL querying over relational databases;
BFCL~\citep{patil2025bfcl}, multi-turn user requests served by calling
stateful APIs; and ScienceWorld~\citep{wang2022scienceworld0},
interactive science experiments in a text environment. Every stream
consists of consecutive blocks of 50 training tasks per group, evaluated
on disjoint hold-out sets $\mathcal{D}_k^{\text{eval}}$ of 50 tasks per
group (90 for ScienceWorld). Following \S\ref{sec:streams}, we construct
four drifting streams that isolate three types of distribution shift and
their combination:
\begin{itemize}
  \item \emph{Topic drift} ($T = 150, K = 3$). The environment and
  workflow stay fixed while the underlying topic changes: the
  agent trains on three BIRD databases in sequence, \texttt{BIRD-F1}
  (\texttt{formula\_1}) $\to$ \texttt{BIRD-Code}
  (\texttt{codebase\_community}) $\to$ \texttt{BIRD-TP}
  (\texttt{thrombosis\_prediction}). While the high-level text-to-SQL
  task remains the same, each database carries its own schema and local
  conventions that must be learned without transferring stale
  assumptions to the next.
  \item \emph{Workflow drift} ($T = 100, K = 2$). The environment and
  tool catalog stay fixed while the workflow required for a correct
  response changes: the agent trains on \texttt{BFCL-Miss} followed by
  \texttt{BFCL-Base}. Both groups use the same APIs, but
  \texttt{BFCL-Miss} requires the agent to ask the user to clarify
  missing arguments, whereas \texttt{BFCL-Base} supplies all required
  arguments and penalizes unnecessary clarification.
  \item \emph{Domain drift} ($T = 150, K = 3$). The environment itself
  changes along with its observation and action spaces: the agent trains
  across ScienceWorld $\to$ \texttt{BFCL-Base} $\to$ \texttt{BIRD-F1},
  spanning interactive embodied environments, multi-turn tool use, and
  database querying.
  \item \emph{Compound drift} ($T = 200, K = 4$). Over a longer
  deployment horizon, shifts in topic, domain, and workflow occur in
  sequence rather than in isolation: \texttt{BIRD-Code} $\to$
  \texttt{BIRD-F1} (\emph{topic}) $\to$ \texttt{BFCL-Miss}
  (\emph{domain}) $\to$ \texttt{BFCL-Base} (\emph{workflow}), requiring
  the evolver to handle both cross-domain transitions and conflicting
  within-domain shifts in a single stream.
\end{itemize}

\paragraph{Model and harness.}
We hold the base model fixed across methods and streams so that
differences arise from the evolver alone, and use Gemini 3.6
Flash~\citep{gemini36flash} in a ReAct-style loop~\citep{yao2023react},
which alternates environment observations with actions. Per-family
action interfaces are described in Appendix~\ref{app:impl}.

\begin{table}[t]
  \centering
  \caption{Continual learning performance across the four drift scenarios
  on Gemini 3.6 Flash. Each stream is listed by its ordered task groups,
  followed by the per-stream \emph{Average} and the four-stream
  \textbf{Overall average}.
  \textbf{Seed} reports baseline accuracy $B_k$ (\%); method columns
  report terminal advantage ($\text{Term.}$), average advantage
  ($\bar{\Delta}$, Eq.~\ref{eq:avg-advantage}), within-block plasticity
  ($P$, Eq.~\ref{eq:plasticity}), and cross-block stability
  ($S = -(I_k + F_k)$, Eq.~\ref{eq:stability}), all in percentage points
  (pp.) over $B_k$. Summary rows are shaded green for gains and red for
  losses, with the best method per metric in bold; the separate $-I_k$
  and $-F_k$ decomposition is in Table~\ref{tab:group-decomp}
  (Appendix~\ref{app:gepa-occupancy}).}
  \label{tab:main}
  \providecolor{tblgood}{HTML}{15A06F}
\providecolor{tblbad}{HTML}{C7402D}
{\footnotesize
\renewcommand{\arraystretch}{1.12}
\setlength{\tabcolsep}{0pt}
\resizebox{\linewidth}{!}{%
\begin{tabular}{l@{\hspace{6pt}}l@{\hspace{8pt}}r@{\hspace{11pt}}>{\hspace{2.5pt}}r<{\hspace{2.5pt}}@{\hspace{1.5pt}}>{\hspace{2.5pt}}r<{\hspace{2.5pt}}@{\hspace{5pt}}>{\hspace{2.5pt}}r<{\hspace{2.5pt}}@{\hspace{1.5pt}}>{\hspace{2.5pt}}r<{\hspace{2.5pt}}@{\hspace{11pt}}>{\hspace{2.5pt}}r<{\hspace{2.5pt}}@{\hspace{1.5pt}}>{\hspace{2.5pt}}r<{\hspace{2.5pt}}@{\hspace{5pt}}>{\hspace{2.5pt}}r<{\hspace{2.5pt}}@{\hspace{1.5pt}}>{\hspace{2.5pt}}r<{\hspace{2.5pt}}@{\hspace{11pt}}>{\hspace{2.5pt}}r<{\hspace{2.5pt}}@{\hspace{1.5pt}}>{\hspace{2.5pt}}r<{\hspace{2.5pt}}@{\hspace{5pt}}>{\hspace{2.5pt}}r<{\hspace{2.5pt}}@{\hspace{1.5pt}}>{\hspace{2.5pt}}r<{\hspace{2.5pt}}}
\toprule
&& & \multicolumn{4}{c}{ReasoningBank} & \multicolumn{4}{c}{GEPA} & \multicolumn{4}{c}{Meta-Harness} \\
\cmidrule(lr){4-7} \cmidrule(lr){8-11} \cmidrule(lr){12-15}
Stream & Task group & Seed & Term. & $\bar{\Delta}$ & $P$ & $S$ & Term. & $\bar{\Delta}$ & $P$ & $S$ & Term. & $\bar{\Delta}$ & $P$ & $S$ \\
\midrule
Topic & BIRD-F1 & $52.5$ & \textcolor{black!45}{$0.0$} & \textcolor{black!45}{$0.5$} & $1.5$ & $-1.5$ & $-2.0$ & $3.0$ & $7.5$ & $-9.5$ & $-3.5$ & $-4.5$ & $-4.5$ & \textcolor{black!45}{$0.0$} \\
 & BIRD-Code & $61.8$ & \textcolor{black!45}{$-0.5$} & \textcolor{black!45}{$-0.2$} & $-1.0$ & \textcolor{black!45}{$0.0$} & \textcolor{black!45}{$0.5$} & \textcolor{black!45}{$0.0$} & $1.0$ & $-2.0$ & $1.5$ & $2.0$ & $1.0$ & \textcolor{black!45}{$0.0$} \\
 & BIRD-TP & $43.2$ & $1.5$ & \textcolor{black!45}{$-0.7$} & $1.5$ & $-1.8$ & $-3.5$ & $-3.2$ & $-3.5$ & $-3.0$ & $-2.5$ & $-1.5$ & $-2.5$ & $-1.0$ \\
\addlinespace[1pt]
 & \textit{Average} & $52.5$ & \cellcolor{tblgood!8}$\mathbf{0.3}$ & \cellcolor{tblbad!8}$-0.1$ & \cellcolor{tblgood!9}$0.7$ & \cellcolor{tblbad!10}$-1.1$ & \cellcolor{tblbad!12}$-1.7$ & \cellcolor{tblbad!8}$-0.1$ & \cellcolor{tblgood!12}$\mathbf{1.7}$ & \cellcolor{tblbad!20}$-4.8$ & \cellcolor{tblbad!11}$-1.5$ & \cellcolor{tblbad!11}$-1.3$ & \cellcolor{tblbad!13}$-2.0$ & \cellcolor{tblbad!8}$\mathbf{-0.3}$ \\
\midrule
Workflow & BFCL-Miss & $53.8$ & $4.5$ & $8.5$ & $12.5$ & $-8.0$ & $2.5$ & $9.0$ & $15.5$ & $-13.0$ & $11.5$ & $10.5$ & $9.5$ & \textcolor{black!45}{$0.0$} \\
 & BFCL-Base & $77.3$ & $4.5$ & $1.5$ & $4.5$ & $-1.5$ & $6.5$ & $3.9$ & $6.5$ & \textcolor{black!45}{$0.0$} & $-2.0$ & $-1.5$ & $-2.0$ & $-1.0$ \\
\addlinespace[1pt]
 & \textit{Average} & $65.5$ & \cellcolor{tblgood!19}$4.5$ & \cellcolor{tblgood!20}$5.0$ & \cellcolor{tblgood!28}$8.5$ & \cellcolor{tblbad!20}$-4.8$ & \cellcolor{tblgood!19}$4.5$ & \cellcolor{tblgood!24}$\mathbf{6.4}$ & \cellcolor{tblgood!28}$\mathbf{11.0}$ & \cellcolor{tblbad!24}$-6.5$ & \cellcolor{tblgood!20}$\mathbf{4.8}$ & \cellcolor{tblgood!19}$4.5$ & \cellcolor{tblgood!17}$3.8$ & \cellcolor{tblbad!9}$\mathbf{-0.5}$ \\
\midrule
Domain & ScienceWorld & $78.9$ & $5.3$ & $3.0$ & $1.9$ & \textcolor{black!45}{$0.0$} & $7.5$ & $7.2$ & $4.7$ & \textcolor{black!45}{$0.0$} & $4.4$ & $7.2$ & $8.9$ & $-4.4$ \\
 & BFCL-Base & $79.2$ & $1.5$ & \textcolor{black!45}{$0.2$} & \textcolor{black!45}{$0.2$} & $-1.2$ & $5.0$ & \textcolor{black!45}{$0.9$} & $3.2$ & $-5.5$ & $-2.2$ & \textcolor{black!45}{$-0.8$} & $1.2$ & $-5.0$ \\
 & BIRD-F1 & $50.8$ & $1.0$ & $1.2$ & $1.0$ & \textcolor{black!45}{$0.0$} & $6.0$ & \textcolor{black!45}{$0.8$} & $6.0$ & $-1.8$ & $2.5$ & $5.0$ & $2.5$ & \textcolor{black!45}{$0.0$} \\
\addlinespace[1pt]
 & \textit{Average} & $69.6$ & \cellcolor{tblgood!14}$2.6$ & \cellcolor{tblgood!11}$1.4$ & \cellcolor{tblgood!10}$1.1$ & \cellcolor{tblbad!9}$\mathbf{-0.4}$ & \cellcolor{tblgood!23}$\mathbf{6.2}$ & \cellcolor{tblgood!15}$3.0$ & \cellcolor{tblgood!19}$\mathbf{4.7}$ & \cellcolor{tblbad!14}$-2.4$ & \cellcolor{tblgood!12}$1.6$ & \cellcolor{tblgood!17}$\mathbf{3.8}$ & \cellcolor{tblgood!18}$4.2$ & \cellcolor{tblbad!15}$-3.1$ \\
\midrule
Compound & BIRD-Code & $61.5$ & $-1.0$ & \textcolor{black!45}{$0.6$} & $1.0$ & $-2.0$ & $6.0$ & $5.0$ & $7.0$ & $-1.0$ & $1.5$ & $1.0$ & $1.5$ & \textcolor{black!45}{$0.0$} \\
 & BIRD-F1 & $50.7$ & $4.5$ & $2.8$ & $2.5$ & \textcolor{black!45}{$0.0$} & \textcolor{black!45}{$0.0$} & $1.8$ & $4.0$ & $-4.0$ & \textcolor{black!45}{$-0.5$} & $-1.5$ & $-2.0$ & $-1.5$ \\
 & BFCL-Miss & $52.3$ & $3.8$ & $1.1$ & $5.0$ & $-3.5$ & $12.0$ & $3.4$ & $14.0$ & $-8.1$ & $3.5$ & $1.8$ & $13.8$ & $-15.2$ \\
 & BFCL-Base & $77.1$ & $3.8$ & $1.7$ & $3.8$ & \textcolor{black!45}{$0.0$} & $5.0$ & $2.2$ & $5.0$ & \textcolor{black!45}{$-0.2$} & $-1.5$ & \textcolor{black!45}{$-0.6$} & $-1.5$ & \textcolor{black!45}{$-0.7$} \\
\addlinespace[1pt]
 & \textit{Average} & $60.4$ & \cellcolor{tblgood!15}$2.8$ & \cellcolor{tblgood!11}$1.5$ & \cellcolor{tblgood!15}$3.1$ & \cellcolor{tblbad!11}$\mathbf{-1.4}$ & \cellcolor{tblgood!22}$\mathbf{5.8}$ & \cellcolor{tblgood!15}$\mathbf{3.1}$ & \cellcolor{tblgood!26}$\mathbf{7.5}$ & \cellcolor{tblbad!16}$-3.3$ & \cellcolor{tblgood!10}$0.8$ & \cellcolor{tblgood!8}$0.2$ & \cellcolor{tblgood!15}$2.9$ & \cellcolor{tblbad!19}$-4.4$ \\
\midrule
\multicolumn{2}{l}{\textbf{Overall average}} & $62.0$ & \cellcolor{tblgood!18}$2.5$ & \cellcolor{tblgood!16}$2.0$ & \cellcolor{tblgood!21}$3.3$ & \cellcolor{tblbad!16}$\mathbf{-1.9}$ & \cellcolor{tblgood!22}$\mathbf{3.7}$ & \cellcolor{tblgood!20}$\mathbf{3.1}$ & \cellcolor{tblgood!30}$\mathbf{6.2}$ & \cellcolor{tblbad!24}$-4.3$ & \cellcolor{tblgood!14}$1.4$ & \cellcolor{tblgood!16}$1.8$ & \cellcolor{tblgood!17}$2.2$ & \cellcolor{tblbad!17}$-2.1$ \\
\bottomrule
\end{tabular}%
}\par}

\end{table}

\paragraph{Self-evolving algorithms.}
We compare three representative methods that span experience
augmentation and self-modification across progressively larger update
spaces: ReasoningBank~\citep{ouyang2025reasoningbank0}, which distills
completed episodes into memory items in an external store and retrieves
the top-$1$ most similar item into the context of later episodes;
GEPA~\citep{agrawal2025gepa}, which rewrites a shared strategy section
in the system prompt by reflective mutation over collected rollouts; and
Meta-Harness~\citep{lee2026meta0harness0}, which proposes edits to a
persistent workspace (including the system prompt, memory and skill
files, and the controller code \texttt{harness.py}) through a coding
agent and accepts an edit only when it improves validation accuracy on
the active sub-block. Every run starts from the same seed harness $c_0$
with an empty state and receives binary task-success feedback
$r_t \in \{0, 1\}$ after each training task.

\paragraph{Evaluation protocol.}
We checkpoint the harness at every block boundary (every 50 stream
positions) and evaluate each frozen checkpoint on the hold-out sets of
all $K$ groups in the stream, averaging over eight rollouts per instance
to estimate $A[t, k]$. Further details are in Appendix~\ref{app:impl}.

\begin{figure}[t]
  \centering
  \includegraphics[width=\linewidth]{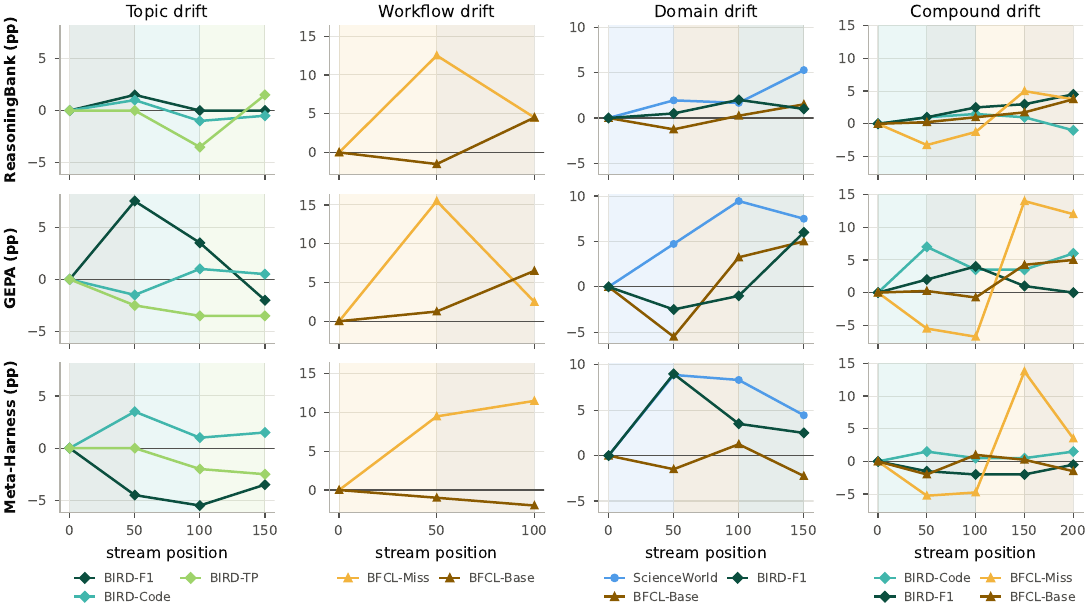}
  \caption{Per-group learning dynamics across the four drift scenarios
  (columns) and three methods (rows), plotted as the advantage over each
  method's seed harness, $A[t, k] - B_k$ (pp.). Each group has one
  color, used for its curve and for the wash of the block in which it is
  trained.}
  \label{fig:drifting-curves}
\end{figure}

\subsection{Current Methods Trade Stability for Plasticity}
\label{sec:main-results}

Evaluating the three methods across the four drifting streams shows that
self-evolution improves both terminal and average performance in most
settings (Table~\ref{tab:main}; Figure~\ref{fig:drifting-curves}).
All three methods achieve positive gains on workflow, domain, and
compound drift, as well as in the mean across the four scenarios:
averaged over streams, terminal advantage reaches $+3.7$ for GEPA,
$+2.5$ for ReasoningBank, and $+1.4$ for Meta-Harness, with average
advantage $\bar{\Delta}$ between $+1.8$ and $+3.1$. Only topic drift
yields near-zero or slightly negative net change ($-1.7$ for GEPA,
$+0.3$ for ReasoningBank, and $-1.5$ for Meta-Harness).

\begin{figure}[t]
  \centering
  \includegraphics[width=\linewidth]{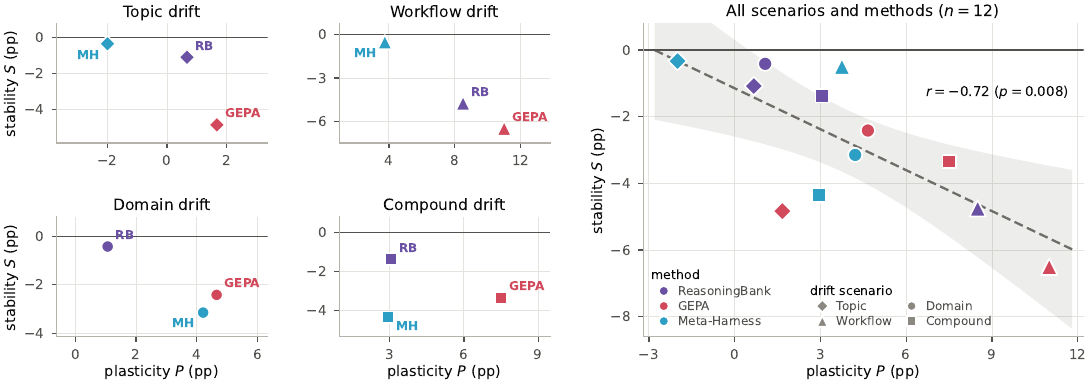}
  \caption{Plasticity $P$ against stability $S$ across the four drift
  scenarios. \textbf{Left:} per-scenario scatter (one point per method).
  \textbf{Right:} overall correlation across all four scenarios and three
  methods ($n = 12$; dashed line and shaded band show the linear
  least-squares fit and $95\%$ confidence interval).}
  \label{fig:ps-frontier}
\end{figure}

Decomposing these aggregate scores into within-block plasticity $P$ and
cross-block stability $S$, however, reveals a consistent negative
correlation between learning the active block and preserving performance
on the others (Table~\ref{tab:main}; Figure~\ref{fig:ps-frontier}).
Across all four scenarios, plasticity and stability exhibit a clear
trade-off: higher plasticity on the group being trained is accompanied by
larger stability losses across blocks, so the most plastic method in each
scenario is always less stable than the most conservative one.

This trade-off reflects where each method applies its updates. GEPA
rewrites a single shared strategy prompt that is injected into every
rollout. A rule discovered during a block therefore changes the agent's
behavior broadly across all tasks of that group, giving GEPA the highest
plasticity in every scenario ($+6.2$ on average, reaching $+11.0$ on
workflow drift and $+7.5$ on compound drift). Because later blocks
rewrite the same shared prompt and apply it to every group, however,
GEPA also incurs the largest stability loss ($-4.3$ on average).
ReasoningBank, by contrast, leaves the system prompt untouched,
appending one memory item per episode to an external store and
retrieving only the single most similar item at read time. Because an
item is injected only into episodes whose query resembles its source
episode, updates remain localized and ReasoningBank achieves the highest
average stability ($-1.9$, with $-0.4$ on domain drift and $-1.4$ on
compound drift), at the cost of roughly half the plasticity of GEPA
($+3.3$ on average). Like GEPA, Meta-Harness modifies a shared global
workspace and is the least stable method on domain and compound drift
($S = -3.1$ and $-4.4$). It appears stable on topic and workflow drift
($S = -0.3$ and $-0.5$) only because its validation gate rejects half or
more of the candidate updates, freezing the workspace on later blocks at
the cost of low plasticity. Removing this gate
so that every checkpoint accepts the best candidate drops its stability
to $-3.1$ on topic drift and $-5.5$ on workflow drift.

\subsection{Self-Modification Causes Cross-Task Interference and Overwriting}
\label{sec:transfer}

\paragraph{Pre-block interference from task-specific rules.}
In self-modifying methods (GEPA and Meta-\allowbreak Harness), every update writes to
a shared instruction or workspace surface that is injected into all
subsequent rollouts, creating two forms of instability: pre-block
interference $I_k$ and post-block forgetting $F_k$
(Eq.~\ref{eq:interf-forget}). Before a group's training block begins
($t \le \sigma_k$), rules specialized to earlier blocks are applied
unconditionally to unseen tasks, pushing their accuracy below the seed
harness $B_k$ (Table~\ref{tab:main} and Figure~\ref{fig:drifting-curves};
full $-I_k$ and $-F_k$ breakdown in Table~\ref{tab:group-decomp},
Appendix~\ref{app:gepa-occupancy}). Across unrelated domains, training on
ScienceWorld in domain drift lowers \texttt{BFCL-Base} at position 50 by
$5.5$ under GEPA and $1.5$ under Meta-Harness, and training on the two
BIRD blocks in compound drift degrades \texttt{BFCL-Miss} before a single
BFCL task has arrived, reaching $-6.8$ at position 100 under GEPA and
$-4.8$ under Meta-Harness. Inspecting the evolved prompts at position 100
of compound drift explains this drop
(Figure~\ref{fig:interference-prompts} in
Appendix~\ref{app:gepa-occupancy}): after the two BIRD blocks, both
methods embed mandatory SQLite exploration and interactive verification
instructions directly into the global system prompt, which biases the
agent on incoming \texttt{BFCL-Miss} tasks to invoke exploratory lookup
tools rather than ask the user for missing arguments. Even when tasks
are closely related and differ only in topic, as in topic drift, we
still observe such interference: adapting to \texttt{BIRD-F1} and
\texttt{BIRD-Code} lowers \texttt{BIRD-TP} at
position 100 by $3.5$ under GEPA and $2.0$ under Meta-Harness.

\begin{figure}[t]
  \centering
  \includegraphics[width=\linewidth]{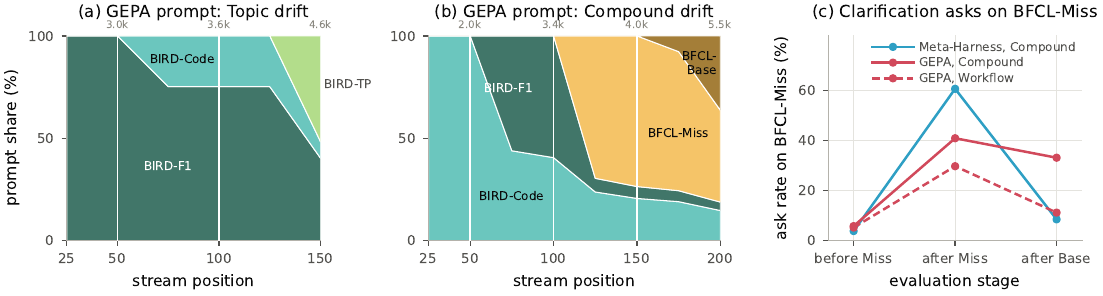}
  \caption{Mechanisms of post-block forgetting under self-modification.
  \textbf{(a,b)} Share of GEPA's strategy prompt characters attributed to
  the block in which each line was first written on topic drift
  \textbf{(a)} and compound drift \textbf{(b)} (numbers above each panel
  give total strategy length; full four-stream comparison in
  Figure~\ref{fig:gepa-occupancy}, Appendix~\ref{app:gepa-occupancy}).
  \textbf{(c)} Percentage of \texttt{BFCL-Miss} evaluation episodes in
  which the agent withholds tool calls and asks for clarification before
  the \texttt{BFCL-Miss} block, after the \texttt{BFCL-Miss} block, and
  after the subsequent \texttt{BFCL-Base} block.}
  \label{fig:mechanisms}
\end{figure}

\paragraph{Post-block forgetting via compression and overwriting.}
After a group's training block ends ($t > \tau_k$), later blocks
continue to modify the same shared prompt or workspace, compressing or
overwriting earlier instructions and causing substantial post-block
forgetting (Table~\ref{tab:main} and Table~\ref{tab:group-decomp} in
Appendix~\ref{app:gepa-occupancy}). Tracking the character share
of each group's rules in GEPA's strategy prompt across checkpoints makes
this displacement directly visible (Figure~\ref{fig:mechanisms}a,b; see
Figure~\ref{fig:gepa-occupancy} in Appendix~\ref{app:gepa-occupancy} for
all four streams). On topic drift (Figure~\ref{fig:mechanisms}a), rules
written during the \texttt{BIRD-F1} block ($+7.5$ advantage at
position 50) fall from $100\%$ of the prompt at position 50 to $75.2\%$
during \texttt{BIRD-Code} (position 100) and $40.2\%$ during
\texttt{BIRD-TP} (position 150, where
\texttt{BIRD-TP} takes over $51.7\%$ of the prompt), and
\texttt{BIRD-F1}'s accuracy drops in lockstep to $+3.5$ at position
100 and $-2.0$ at position 150. Similarly, on compound drift
(Figure~\ref{fig:mechanisms}b), \texttt{BIRD-Code}'s share
falls from $100\%$ at position 50 to $40.5\%$ at position 100 and
$14.7\%$ at position 200; when the stream transitions from BIRD to
\texttt{BFCL-Miss} after position 100, GEPA condenses the seven BIRD SQL
sections ($3{,}386$ characters at position 100) into a single
$1{,}034$-character section, reducing \texttt{BIRD-F1}'s share from
$59.5\%$ at position 100 to $5.8\%$ at position 150 ($4.1\%$ at position
200) and erasing its $+4.0$ gain by the end of the stream. Even when all
four groups are interleaved uniformly rather than arriving in blocks,
competing updates continue to contend for space in the shared prompt and
degrade GEPA's terminal advantage from $+5.8$ to $+0.9$
(Appendix~\ref{app:mix}).

\paragraph{Spontaneous additive updates across distant domains are fragile.}
Occasionally, when consecutive blocks come from distant domains that
share neither tools nor observations, the evolver spontaneously adopts
an additive update pattern that prevents forgetting. On domain drift,
GEPA's strategy prompt grows from $4{,}372$ characters after
ScienceWorld (position 50) to $5{,}650$ after \texttt{BFCL-Base}
(position 100) and $7{,}621$ after \texttt{BIRD-F1} (position 150), appending new
domain sections while retaining all six ScienceWorld sections verbatim
(Figure~\ref{fig:additive-prompt}, left, in
Appendix~\ref{app:gepa-occupancy}) and achieving $\bar{F} = 0.0$
(Table~\ref{tab:main} and Table~\ref{tab:group-decomp}). This additive behavior, however, emerges
unreliably: on compound drift, when the stream transitions across
domains from BIRD to \texttt{BFCL-Miss} after position 100, GEPA instead
compresses the seven BIRD SQL sections into a single generic section
(Figure~\ref{fig:additive-prompt}, right), causing the \texttt{BIRD-F1}
advantage to drop from $+4.0$ at position 100 to $0.0$ at position 200
(Table~\ref{tab:main} and Figure~\ref{fig:drifting-curves}).

\paragraph{Behavioral reversal under conflicting workflows.}
Forgetting is most severe when consecutive blocks have nearly identical
queries and domains but require opposite workflows, as in workflow drift
and the BFCL transition of compound drift
(Figure~\ref{fig:mechanisms}c). Because the surface domain and tool
catalog do not change, the model fails to recognize a shift and rewrites
the earlier policy into its opposite. Before the \texttt{BFCL-Miss}
block, the seed agent calls tools speculatively on $91\%$ to $93\%$ of
\texttt{BFCL-Miss} episodes and asks the user for missing arguments on
only $3.8\%$ to $5.8\%$. Training on \texttt{BFCL-Miss} teaches both
self-modifying methods to withhold speculative calls and ask for
clarification, raising the clarification rate after the
\texttt{BFCL-Miss} block to $29.8\%$ under GEPA on workflow drift
($+15.5$ accuracy gain) and $60.8\%$ under Meta-Harness on compound
drift ($+13.8$). In the subsequent \texttt{BFCL-Base} block, however,
all required arguments are already supplied and asking unnecessary
clarifying questions is penalized. Both methods consequently rewrite
their instructions to execute tool calls directly without asking the
user. On workflow drift, this rewrite under GEPA cuts the clarification
rate on \texttt{BFCL-Miss} from $29.8\%$ ($119/400$ rollouts) to
$11.2\%$ ($45/400$) and drops \texttt{BFCL-Miss} accuracy by $13.0$
(from $+15.5$ at position 50 to $+2.5$ at position 100;
Table~\ref{tab:main}). Meta-Harness shows the same reversal on
compound drift. At position 150, it writes a rule across its prompt,
memory, and skill files that forbids speculative calls when arguments are
missing. During the \texttt{BFCL-Base} block at position 175, however, an
accepted edit overwrites all three files with instructions to execute
tools immediately in sequence. This reduces clarification asks on
\texttt{BFCL-Miss} from $60.8\%$ to $8.5\%$ ($34/400$) and drops accuracy
by $10.2$ at position 200.

\subsection{Similarity-Based Retrieval Cannot Prevent Interference or Collision}
\label{sec:forgetting}

\begin{figure}[t]
  \centering
  \includegraphics[width=0.95\linewidth]{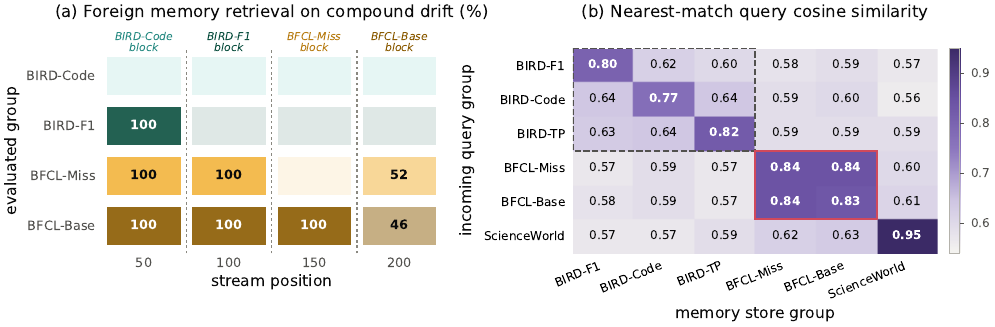}
  \caption{Mechanisms of interference and collision under ReasoningBank's
  similarity-based retrieval. \textbf{(a)} Share (\%) of evaluation
  episodes on compound drift whose retrieved memory came from a foreign
  task group (each row tinted in the evaluated group's color, darker with
  the foreign share; empty cells are $0\%$; dashed lines separate training
  blocks). Before a group's block begins, $100\%$ of retrievals come from
  earlier blocks; at position 200, $52\%$ of \texttt{BFCL-Miss} and $46\%$
  of \texttt{BFCL-Base} episodes retrieve a memory from the opposite BFCL
  workflow. \textbf{(b)} Mean nearest-match query cosine similarity across
  the six task groups (dashed box: BIRD databases; red box:
  \texttt{BFCL-Miss} and \texttt{BFCL-Base}, whose cross-group similarity
  matches their within-group similarity).}
  \label{fig:rb-mechanisms}
\end{figure}

\paragraph{Domain separation prevents post-block forgetting.}
Storing experiences in an external memory bank and retrieving the top
match by query similarity allows ReasoningBank to leave the system
prompt untouched, which eliminates post-block forgetting when task
domains are well separated: on domain drift, queries from ScienceWorld,
\texttt{BFCL-Base}, and BIRD have much higher within-group than
cross-group nearest-match cosine similarity ($0.80$--$0.95$ vs.\
$0.57$--$0.63$; Figure~\ref{fig:rb-mechanisms}b and
Table~\ref{tab:rb-similarity} in Appendix~\ref{app:rb-retrieval}), so
no evaluation episodes ($0\%$) after a block is populated
retrieve a memory from another domain and $\bar{F} = 0.0$. Across the
four streams, however, retrieval alone still fails to prevent continual
learning instability (Table~\ref{tab:main} and
Figure~\ref{fig:rb-mechanisms}).

\paragraph{Pre-block interference from foreign memory retrieval.}
Because retrieval always injects the nearest stored match
without a similarity threshold, any group evaluated before its own
training block begins ($t \le \sigma_k$) has zero in-group entries in
the store and receives foreign memories in $100\%$ of episodes
(Figure~\ref{fig:rb-mechanisms}a). On domain drift, injecting
ScienceWorld memories into \texttt{BFCL-Base} at position 50 lowers
accuracy by $1.2$; on compound drift, every \texttt{BFCL-Miss} episode
at positions 50 and 100 receives a BIRD SQL reflection, pulling
\texttt{BFCL-Miss} up to $3.2$ below the seed; and on topic drift,
retrieving \texttt{BIRD-F1} and \texttt{BIRD-Code} memories
lowers \texttt{BIRD-TP} by $3.5$ at position 100. While
adding a similarity threshold could mitigate this unconditional
retrieval, query similarity alone struggles to distinguish how related
two queries actually are: even within the same text-to-SQL task, queries
across different topics have a nearest-match similarity ($0.60$--$0.64$,
dashed box in Figure~\ref{fig:rb-mechanisms}b) that is barely higher than
the similarity across completely different tasks and domains
($0.56$--$0.63$).

\paragraph{Retrieval collision when similar queries require conflicting workflows.}
Even with a similarity threshold, retrieval cannot prevent
pre-block interference or post-block forgetting when distinct task groups
occupy the same region of the embedding space
(Figure~\ref{fig:rb-mechanisms}b). Under workflow shifts,
\texttt{BFCL-Miss} and \texttt{BFCL-Base} queries refer to the same API
catalog and are phrased in nearly identical language, exhibiting a mean
pairwise cosine similarity of $0.667$ and a mean nearest-match similarity
of $0.835$--$0.840$, which is indistinguishable from their within-group
similarity ($0.664$--$0.666$ pairwise and $0.834$--$0.837$ nearest-match;
red box in Figure~\ref{fig:rb-mechanisms}b). No similarity threshold can separate
the two groups. Before the \texttt{BFCL-Base} block begins on workflow
drift, \texttt{BFCL-Miss} abstention memories are retrieved into
\texttt{BFCL-Base} episodes, causing pre-block interference ($-1.5$ at
position 50). Conversely, once the \texttt{BFCL-Base} block populates
the store with memories that reward multi-step tool execution, $52\%$ of
\texttt{BFCL-Miss} evaluation episodes at the end of both workflow drift
and compound drift (Figure~\ref{fig:rb-mechanisms}a) retrieve a
\texttt{BFCL-Base} memory (while $46\%$ of \texttt{BFCL-Base} episodes
retrieve a \texttt{BFCL-Miss} memory), displacing the earlier abstention
examples and dropping \texttt{BFCL-Miss} accuracy by $8.0$ on workflow
drift ($+12.5 \to +4.5$). Semantic similarity between queries captures
topical overlap, but cannot determine whether a stored trajectory's
workflow is compatible with the incoming task.

\section{Related Work}
\label{sec:related}

\paragraph{Self-evolving agents.}
Self-evolving agents update themselves from their own experience rather
than remaining fixed after training. In the nonparametric form studied
here, the evolver edits the harness and leaves the weights untouched,
following one of two paradigms. \emph{Self-modification} rewrites the
artifacts the agent runs on: automatic prompt optimization such as
GEPA~\citep{agrawal2025gepa}, TextGrad~\citep{yuksekgonul2024textgrad},
and PromptBreeder~\citep{fernando2024promptbreeder} rewrites the system
prompt from rollout feedback, while ADAS~\citep{hu2024adas}, Darwin
G\"odel Machine~\citep{zhang2025darwin}, EvoTest~\citep{he2026evotest},
and Meta-Harness~\citep{lee2026meta0harness0} extend the edits to the
agent's code~\citep{yin2024g0del}. \emph{Experience augmentation} keeps
the artifacts fixed and accumulates an external store that later
episodes retrieve from: reflections in Reflexion~\citep{shinn2023reflexion}
and ExpeL~\citep{zhao2024expel}, skills in Voyager~\citep{wang2023voyager},
and memory items in ReasoningBank~\citep{ouyang2025reasoningbank0},
MemGen~\citep{zhang2025memgen0}, and Mem0~\citep{chhikara2025mem0}. Parametric counterparts instead
update the weights at test time, through self-correction training in
SCoRe~\citep{kumar2025score}, reinforcement learning in
TTRL~\citep{zuo2026ttrl} and TTT-Discover~\citep{yuksekgonul2026learning},
or a learned evolver in LSE~\citep{chen2026learning}. Unlike prior work
evaluated within a single task family, we study nonparametric methods on
heterogeneous streams.

\paragraph{Continual learning.}
Continual learning studies a learner that faces a sequence of tasks and
must acquire each without degrading performance on earlier ones, known as
catastrophic forgetting~\citep{mccloskey1989catastrophic,
french1999catastrophic}. Gradient-based methods counter it by protecting
weights important to earlier tasks~\citep{kirkpatrick2017overcoming},
rehearsing stored examples~\citep{lopezpaz2017gradient,
chaudhry2019efficient, buzzega2020dark}, or shaping updates to maximize
transfer and minimize interference~\citep{riemer2018learning}; see
\citet{parisi2019continual} and \citet{delange2021continual} for
surveys. Our plasticity and stability metrics adapt standard
continual-learning evaluation~\citep{lopezpaz2017gradient,
kemker2018measuring}; task-free and online formulations relax explicit
task boundaries~\citep{aljundi2019taskfree, koh2022online}, and continual
evaluation exposes transient drops missed by end-of-task
evaluation~\citep{delange2023stability}, motivating our checkpoint
resolution. For LLMs, continual learning has focused on weight
updates~\citep{shi2024continual, wu2024continual}: continual pre-training
on emerging corpora~\citep{ke2023continual, jang2022towards}, forgetting
under sequential fine-tuning~\citep{luo2023empirical}, and mitigation
algorithms~\citep{shenfeld2026self0distillation}. Rather than updating
weights, the agents studied here keep weights frozen and carry their
state in the harness.

\section{Conclusion}
\label{sec:conclusion}

We asked whether nonparametric self-evolving LLM agents are effective
continual learners, and found that current methods remain caught in a
trade-off between plasticity and stability. Across four drifting stream
scenarios spanning shifts in topic, workflow, domain, and their
combination, self-evolution improves overall performance over the seed
harness in most settings, yet no method learns new task groups strongly
without degrading others. Self-modifying methods (GEPA and Meta-Harness)
can achieve high plasticity when rewriting a shared prompt, but because
task-specific rules are applied globally across all subsequent rollouts,
instructions learned on one block interfere with unseen tasks and are
compressed or overwritten by later blocks, most severely when consecutive
blocks share a domain but require conflicting workflows. Experience
augmentation via query-similarity retrieval (ReasoningBank) avoids
overwriting the shared prompt and prevents forgetting across distant
domains, but still injects foreign memories before a group is
encountered and suffers from retrieval collision when distinct workflows
occupy the same region of the embedding space. Together, these findings
expose the limitations of current self-evolving agents as continual
learners.

\bibliographystyle{abbrvnat}
\bibliography{references}

\newpage
\appendix

\section{Results on the Interleaved Mixture Stream}
\label{app:mix}

As a stationary reference, we also evaluate a \emph{mixture} stream in
which $w_t$ is uniform across $t$ and every position samples a random
task group, corresponding to rehearsal in gradient-based continual
learning~\citep{lopezpaz2017gradient, chaudhry2019efficient}. To test
whether interleaving task groups throughout the stream eliminates the
instability observed on drifting streams, we evaluate a uniformly
interleaved \texttt{Mix} schedule on the four-group compound stream
(\texttt{BIRD-Code}, \texttt{BIRD-F1}, \texttt{BFCL-Miss},
and \texttt{BFCL-Base} sampled uniformly across $T = 200$ tasks;
Table~\ref{tab:mix-compound}). Because all groups are active throughout,
pre-block interference and post-block forgetting are not defined, and we
compare terminal advantage ($\text{Term.}$) and average advantage
($\bar{\Delta}$) directly against the corresponding blocked stream.

\begin{figure}[!ht]
  \centering
  \includegraphics[width=0.82\linewidth]{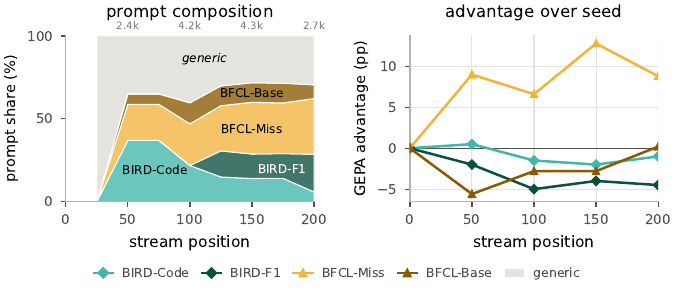}
  \caption{GEPA strategy prompt composition (\textbf{left}) and per-group
  advantage over the seed (\textbf{right}) on the interleaved compound
  \texttt{Mix} stream. Rules not specific to any group are counted as
  generic.}
  \label{fig:gepa-occupancy-mix}
\end{figure}

Interleaving tasks does not resolve the contention on a shared prompt:
GEPA's terminal advantage drops from $+5.8$ on the blocked stream to
$+0.9$ on \texttt{Mix} (with \texttt{BIRD-F1} falling to $-4.5$ and
\texttt{BIRD-Code} to $-1.0$), while Meta-Harness also degrades on
\texttt{BIRD-F1} ($-0.5 \to -3.0$) even as its overall terminal
advantage rises from $+0.8$ to $+1.9$. Figure~\ref{fig:gepa-occupancy-mix}
tracks the composition of GEPA's shared strategy prompt across
checkpoints of the compound \texttt{Mix} stream, where lines are
attributed by benchmark section and group-specific terms (and rules not
specific to any group are counted as generic). Even when all four groups
are active throughout the stream, they contend for space in the single
shared prompt: \texttt{BIRD-Code}'s share falls from $37.0\%$
at position 50 to $21.8\%$ at position 100 and $5.8\%$ at position
200 while \texttt{BIRD-F1} rises from $0\%$ to $22.8\%$, and
competing SQL and BFCL updates leave both BIRD databases below the seed
at the end of the stream.

\begin{table}[!ht]
  \centering
  \caption{Comparison between the blocked drifting schedule and the
  uniformly interleaved \texttt{Mix} schedule on the four-group compound
  stream ($T = 200$), reported in percentage points (pp.) over each
  method's seed harness $B_k$.}
  \label{tab:mix-compound}
  \small
  \setlength{\tabcolsep}{5pt}
  \begin{tabular}{llcccccc}
    \toprule
    \textbf{Method} & \textbf{Schedule} & \textbf{Term.} & $\bar{\Delta}$ & \texttt{BIRD-Code} & \texttt{BIRD-F1} & \texttt{BFCL-Miss} & \texttt{BFCL-Base} \\
    \midrule
    ReasoningBank & Blocked & $+2.8$ & $+1.5$ & $-1.0$ & $+4.5$ & $+3.8$ & $+3.8$ \\
                  & \texttt{Mix} & $+2.3$ & $+1.2$ & $-2.5$ & $+2.0$ & $+7.2$ & $+2.5$ \\
    \midrule
    GEPA          & Blocked & $+5.8$ & $+3.1$ & $+6.0$ & $+0.0$ & $+12.0$ & $+5.0$ \\
                  & \texttt{Mix} & $+0.9$ & $+0.4$ & $-1.0$ & $-4.5$ & $+8.7$ & $+0.2$ \\
    \midrule
    Meta-Harness  & Blocked & $+0.8$ & $+0.2$ & $+1.5$ & $-0.5$ & $+3.5$ & $-1.5$ \\
                  & \texttt{Mix} & $+1.9$ & $+0.5$ & $+1.5$ & $-3.0$ & $+5.2$ & $+3.8$ \\
    \bottomrule
  \end{tabular}
\end{table}

\section{Per-Group Continual Learning Decomposition and Evolved Prompts}
\label{app:gepa-occupancy}

Table~\ref{tab:group-decomp} decomposes each task group's stability
$S_k = -(I_k + F_k)$ from Table~\ref{tab:main} into pre-block
interference $-I_k$ and post-block forgetting $-F_k$ across the four
drifting streams.

\begin{table}[!ht]
  \centering
  \caption{Per-group continual learning decomposition across the four
  drifting streams. $B_k$ is the mean seed accuracy (\%); $P_k$ is
  within-block plasticity at block end $\tau_k$; $-I_k$ is pre-block
  interference prior to $\sigma_k$; $-F_k$ is post-block forgetting from
  $\tau_k$ to $T$; and $\text{Term.}$ is terminal advantage
  $A[T, k] - B_k$ (all in pp.). An em-dash marks a structurally
  undefined cell ($\sigma_k = 0$ for $-I_k$; $\tau_k = T$ for $-F_k$).}
  \label{tab:group-decomp}
  \providecolor{tblgood}{HTML}{15A06F}
\providecolor{tblbad}{HTML}{C7402D}
\providecolor{tblrule}{HTML}{9B9992}
  \scriptsize
  \setlength{\tabcolsep}{1.8pt}
  \resizebox{\linewidth}{!}{%
  \begin{tabular}{llr!{\color{tblrule}\vrule width 0.6pt}rrrr!{\color{tblrule}\vrule width 0.6pt}rrrr!{\color{tblrule}\vrule width 0.6pt}rrrr}
    \toprule
    & & & \multicolumn{4}{c}{ReasoningBank} & \multicolumn{4}{c}{GEPA} & \multicolumn{4}{c}{Meta-Harness} \\
    \cmidrule(lr){4-7} \cmidrule(lr){8-11} \cmidrule(l){12-15}
    Stream & Task group (block) & $B_k$ & $P_k$ & $-I_k$ & $-F_k$ & Term. & $P_k$ & $-I_k$ & $-F_k$ & Term. & $P_k$ & $-I_k$ & $-F_k$ & Term. \\
    \midrule
    Topic & \texttt{BIRD-F1} ($0\!\to\!50$) & $52.5$ & \cellcolor{tblgood!9}$+1.5$ & \textemdash & \cellcolor{tblbad!10}$-1.5$ & $+0.0$ & \cellcolor{tblgood!19}$+7.5$ & \textemdash & \cellcolor{tblbad!27}$-9.5$ & \cellcolor{tblbad!10}$-2.0$ & \cellcolor{tblbad!14}$-4.5$ & \textemdash & $-0.0$ & \cellcolor{tblbad!12}$-3.5$ \\
     & \texttt{BIRD-Code} ($50\!\to\!100$) & $61.8$ & \cellcolor{tblbad!8}$-1.0$ & $-0.0$ & $-0.0$ & \cellcolor{tblbad!7}$-0.5$ & \cellcolor{tblgood!8}$+1.0$ & \cellcolor{tblbad!10}$-1.5$ & \cellcolor{tblbad!8}$-0.5$ & \cellcolor{tblgood!7}$+0.5$ & \cellcolor{tblgood!8}$+1.0$ & $-0.0$ & $-0.0$ & \cellcolor{tblgood!9}$+1.5$ \\
     & \texttt{BIRD-TP} ($100\!\to\!150$) & $43.2$ & \cellcolor{tblgood!9}$+1.5$ & \cellcolor{tblbad!10}$-1.8$ & \textemdash & \cellcolor{tblgood!9}$+1.5$ & \cellcolor{tblbad!12}$-3.5$ & \cellcolor{tblbad!13}$-3.0$ & \textemdash & \cellcolor{tblbad!12}$-3.5$ & \cellcolor{tblbad!11}$-2.5$ & \cellcolor{tblbad!9}$-1.0$ & \textemdash & \cellcolor{tblbad!11}$-2.5$ \\
    \midrule
    Workflow & \texttt{BFCL-Miss} ($0\!\to\!50$) & $53.8$ & \cellcolor{tblgood!27}$+12.5$ & \textemdash & \cellcolor{tblbad!24}$-8.0$ & \cellcolor{tblgood!14}$+4.5$ & \cellcolor{tblgood!32}$+15.5$ & \textemdash & \cellcolor{tblbad!34}$-13.0$ & \cellcolor{tblgood!11}$+2.5$ & \cellcolor{tblgood!22}$+9.5$ & \textemdash & $-0.0$ & \cellcolor{tblgood!25}$+11.5$ \\
     & \texttt{BFCL-Base} ($50\!\to\!100$) & $77.3$ & \cellcolor{tblgood!14}$+4.5$ & \cellcolor{tblbad!10}$-1.5$ & \textemdash & \cellcolor{tblgood!14}$+4.5$ & \cellcolor{tblgood!17}$+6.5$ & $-0.0$ & \textemdash & \cellcolor{tblgood!17}$+6.5$ & \cellcolor{tblbad!10}$-2.0$ & \cellcolor{tblbad!9}$-1.0$ & \textemdash & \cellcolor{tblbad!10}$-2.0$ \\
    \midrule
    Domain & \texttt{ScienceWorld} ($0\!\to\!50$) & $78.9$ & \cellcolor{tblgood!10}$+1.9$ & \textemdash & $-0.0$ & \cellcolor{tblgood!15}$+5.3$ & \cellcolor{tblgood!14}$+4.7$ & \textemdash & $-0.0$ & \cellcolor{tblgood!19}$+7.5$ & \cellcolor{tblgood!21}$+8.9$ & \textemdash & \cellcolor{tblbad!16}$-4.4$ & \cellcolor{tblgood!14}$+4.4$ \\
     & \texttt{BFCL-Base} ($50\!\to\!100$) & $79.2$ & \cellcolor{tblgood!7}$+0.2$ & \cellcolor{tblbad!9}$-1.2$ & $-0.0$ & \cellcolor{tblgood!9}$+1.5$ & \cellcolor{tblgood!12}$+3.2$ & \cellcolor{tblbad!18}$-5.5$ & $-0.0$ & \cellcolor{tblgood!15}$+5.0$ & \cellcolor{tblgood!9}$+1.2$ & \cellcolor{tblbad!10}$-1.5$ & \cellcolor{tblbad!14}$-3.5$ & \cellcolor{tblbad!10}$-2.2$ \\
     & \texttt{BIRD-F1} ($100\!\to\!150$) & $50.8$ & \cellcolor{tblgood!8}$+1.0$ & $-0.0$ & \textemdash & \cellcolor{tblgood!8}$+1.0$ & \cellcolor{tblgood!16}$+6.0$ & \cellcolor{tblbad!10}$-1.8$ & \textemdash & \cellcolor{tblgood!16}$+6.0$ & \cellcolor{tblgood!11}$+2.5$ & $-0.0$ & \textemdash & \cellcolor{tblgood!11}$+2.5$ \\
    \midrule
    Compound & \texttt{BIRD-Code} ($0\!\to\!50$) & $61.5$ & \cellcolor{tblgood!8}$+1.0$ & \textemdash & \cellcolor{tblbad!11}$-2.0$ & \cellcolor{tblbad!8}$-1.0$ & \cellcolor{tblgood!18}$+7.0$ & \textemdash & \cellcolor{tblbad!9}$-1.0$ & \cellcolor{tblgood!16}$+6.0$ & \cellcolor{tblgood!9}$+1.5$ & \textemdash & $-0.0$ & \cellcolor{tblgood!9}$+1.5$ \\
     & \texttt{BIRD-F1} ($50\!\to\!100$) & $50.7$ & \cellcolor{tblgood!11}$+2.5$ & $-0.0$ & $-0.0$ & \cellcolor{tblgood!14}$+4.5$ & \cellcolor{tblgood!13}$+4.0$ & $-0.0$ & \cellcolor{tblbad!15}$-4.0$ & $+0.0$ & \cellcolor{tblbad!10}$-2.0$ & \cellcolor{tblbad!10}$-1.5$ & $-0.0$ & \cellcolor{tblbad!7}$-0.5$ \\
     & \texttt{BFCL-Miss} ($100\!\to\!150$) & $52.3$ & \cellcolor{tblgood!15}$+5.0$ & \cellcolor{tblbad!11}$-2.2$ & \cellcolor{tblbad!9}$-1.2$ & \cellcolor{tblgood!13}$+3.8$ & \cellcolor{tblgood!29}$+14.0$ & \cellcolor{tblbad!20}$-6.1$ & \cellcolor{tblbad!11}$-2.0$ & \cellcolor{tblgood!26}$+12.0$ & \cellcolor{tblgood!29}$+13.8$ & \cellcolor{tblbad!17}$-5.0$ & \cellcolor{tblbad!29}$-10.2$ & \cellcolor{tblgood!12}$+3.5$ \\
     & \texttt{BFCL-Base} ($150\!\to\!200$) & $77.1$ & \cellcolor{tblgood!13}$+3.8$ & $-0.0$ & \textemdash & \cellcolor{tblgood!13}$+3.8$ & \cellcolor{tblgood!15}$+5.0$ & \cellcolor{tblbad!7}$-0.2$ & \textemdash & \cellcolor{tblgood!15}$+5.0$ & \cellcolor{tblbad!9}$-1.5$ & \cellcolor{tblbad!8}$-0.7$ & \textemdash & \cellcolor{tblbad!9}$-1.5$ \\
    \bottomrule
  \end{tabular}%
  }

\end{table}

\begin{figure}[!ht]
  \centering
  \includegraphics[width=\linewidth]{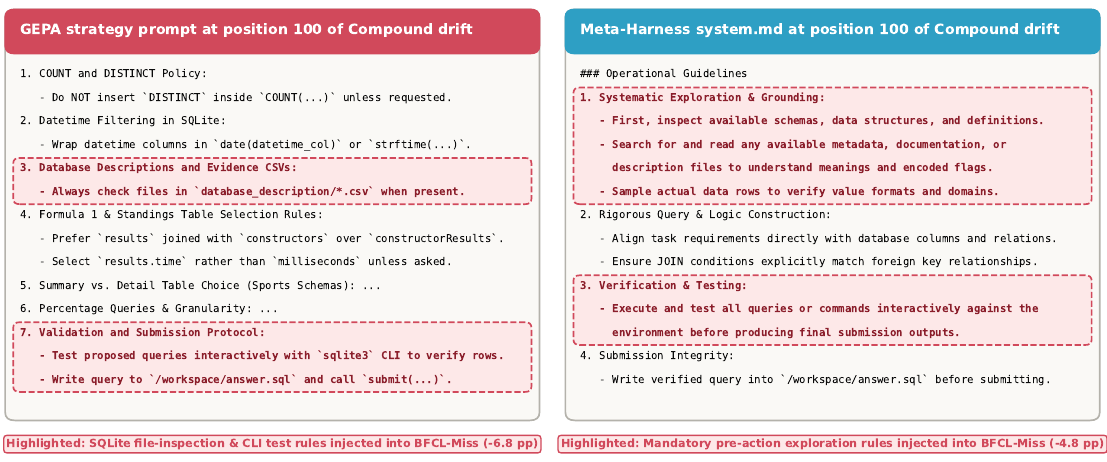}
  \caption{Evolved system prompts at position 100 of compound drift
  (immediately after the two BIRD SQL blocks and before the
  \texttt{BFCL-Miss} block) for GEPA (\textbf{left}) and Meta-Harness
  (\textbf{right}). Highlighted regions mark mandatory SQLite
  exploration, file-inspection, and interactive testing instructions
  that are injected into subsequent \texttt{BFCL-Miss} episodes
  (\S\ref{sec:transfer}), biasing the agent toward calling exploratory
  lookup tools rather than asking the user for missing arguments.}
  \label{fig:interference-prompts}
\end{figure}

\begin{figure}[!ht]
  \centering
  \includegraphics[width=\linewidth]{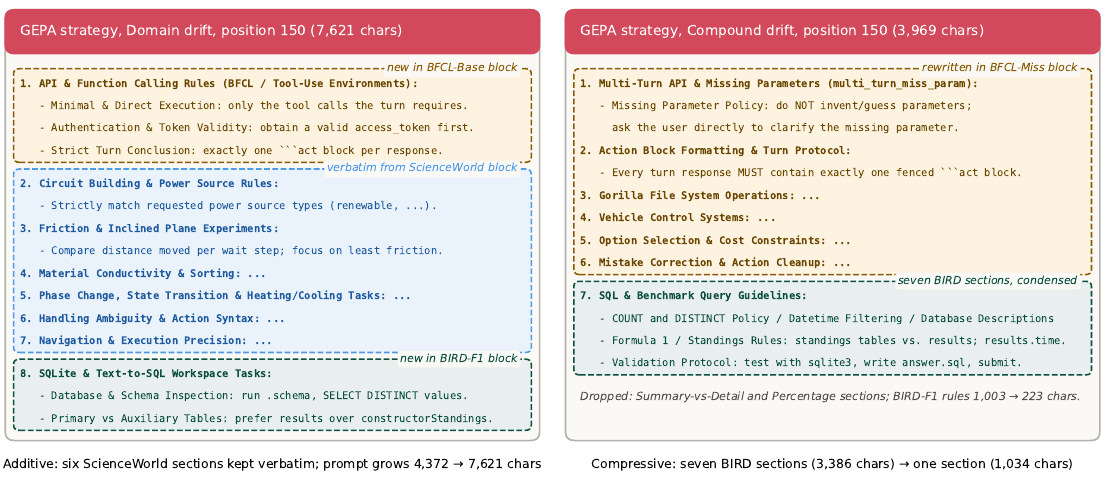}
  \caption{GEPA's evolved strategy at position 150 of domain drift
  (\textbf{left}) and compound drift (\textbf{right}), abridged;
  highlights follow the benchmark palette (BFCL amber, ScienceWorld blue,
  BIRD green) and italic tags are our annotations. On domain drift, GEPA
  appends a \texttt{BFCL-Base} section and a BIRD section while keeping
  all six ScienceWorld sections verbatim ($4{,}372 \to 7{,}621$
  characters). On compound drift, the \texttt{BFCL-Miss} block rewrites
  sections 1--6 and condenses the seven BIRD sections ($3{,}386$
  characters at position 100) into a single $1{,}034$-character section
  7.}
  \label{fig:additive-prompt}
\end{figure}

\begin{figure}[!ht]
  \centering
  \includegraphics[width=\linewidth]{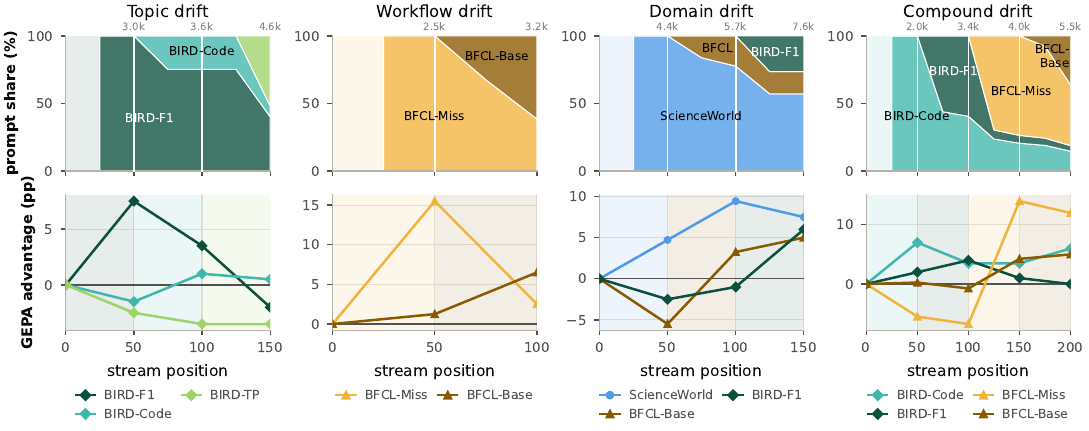}
  \caption{GEPA strategy composition (\textbf{top}; share of strategy
  characters, each line attributed to the block in which it was first
  written, with carried-over or edited lines keeping their original
  group; numbers above each panel give total strategy length) and
  per-group advantage over the seed (\textbf{bottom}) on the four
  drifting streams. Shading marks training blocks.}
  \label{fig:gepa-occupancy}
\end{figure}

Figure~\ref{fig:gepa-occupancy} tracks the composition of GEPA's shared
strategy prompt across checkpoints of the four drifting streams alongside
each group's accuracy advantage over the seed (\S\ref{sec:transfer}). At
each checkpoint, each line in the winning strategy text is attributed to
the block in which it was first written (with carried-over or edited
lines keeping their original group), and each group's share is measured
as the fraction of characters in its lines. On topic, workflow, and
compound drift, later blocks compress or overwrite the sections written
for earlier groups (for example, \texttt{BIRD-F1}'s share falls from
$100\%$ at position 50 to $40.2\%$ at position 150 on topic drift, and
from $59.5\%$ at position 100 to $4.1\%$ at position 200 on compound
drift), and the corresponding accuracy gains decay in lockstep. Only on
domain drift, where ScienceWorld, \texttt{BFCL-Base}, and BIRD have
disjoint action and observation spaces, does GEPA append new sections
without overwriting earlier ones, retaining all six ScienceWorld sections
verbatim ($57.1\%$ of the prompt at position 150).

\section{Similarity and Cross-Group Retrieval under ReasoningBank}
\label{app:rb-retrieval}

ReasoningBank writes one entry to its store per training episode and
injects the most similar stored entry into each evaluation episode
(Appendix~\ref{app:impl}). Table~\ref{tab:rb-similarity} reports both the
mean pairwise cosine similarity and the mean nearest-match cosine
similarity across the six task groups ($50$ training queries per group, $768$-dim
normalized query embeddings; within-group diagonal entries exclude the
trivial self-match $i = j$). Notably, the cross-group similarity between
\texttt{BFCL-Miss} and \texttt{BFCL-Base} ($0.667$ pairwise,
$0.835$--$0.840$ nearest-match) is virtually identical to their
within-group similarity ($0.666$ and $0.664$ pairwise; $0.837$ and
$0.834$ nearest-match), explaining why no similarity threshold can
separate the two workflows. Within BIRD, nearest-match similarity across
different database schemas ($0.601$ to $0.644$) is barely higher than
cross-domain similarity ($0.560$ to $0.626$).

\begin{table}[!ht]
  \centering
  \caption{Query-embedding cosine similarity across the six task groups
  under ReasoningBank ($50$ training tasks per group; diagonal entries exclude
  self-matches). \textbf{Top:} mean pairwise cosine similarity
  $\mathbb{E}_{i \in g_1,\, j \in g_2}[\cos(q_i, q_j)]$.
  \textbf{Bottom:} mean nearest-match cosine similarity
  $\mathbb{E}_{i \in g_1}[\max_{j \in g_2} \cos(q_i, q_j)]$.}
  \label{tab:rb-similarity}
  \small
  \setlength{\tabcolsep}{4pt}
  \begin{tabular}{lcccccc}
    \toprule
    \textbf{Group} & \texttt{BIRD-F1} & \texttt{BIRD-Code} & \texttt{BIRD-TP} & \texttt{BFCL-Miss} & \texttt{BFCL-Base} & \text{ScienceWorld} \\
    \midrule
    \multicolumn{7}{l}{\emph{Mean pairwise cosine similarity}} \\
    \texttt{BIRD-F1}               & \textbf{0.660} & 0.560 & 0.550 & 0.527 & 0.532 & 0.528 \\
    \texttt{BIRD-Code}             & 0.560 & \textbf{0.644} & 0.575 & 0.541 & 0.543 & 0.527 \\
    \texttt{BIRD-TP}               & 0.550 & 0.575 & \textbf{0.707} & 0.536 & 0.535 & 0.545 \\
    \texttt{BFCL-Miss}             & 0.527 & 0.541 & 0.536 & \textbf{0.666} & \textbf{0.667} & 0.568 \\
    \texttt{BFCL-Base}             & 0.532 & 0.543 & 0.535 & \textbf{0.667} & \textbf{0.664} & 0.570 \\
    ScienceWorld                   & 0.528 & 0.527 & 0.545 & 0.568 & 0.570 & \textbf{0.725} \\
    \midrule
    \multicolumn{7}{l}{\emph{Mean nearest-match cosine similarity (row queries against column store)}} \\
    \texttt{BIRD-F1}               & \textbf{0.795} & 0.621 & 0.601 & 0.583 & 0.589 & 0.565 \\
    \texttt{BIRD-Code}             & 0.643 & \textbf{0.770} & 0.638 & 0.589 & 0.600 & 0.560 \\
    \texttt{BIRD-TP}               & 0.628 & 0.644 & \textbf{0.815} & 0.587 & 0.589 & 0.589 \\
    \texttt{BFCL-Miss}             & 0.571 & 0.590 & 0.574 & \textbf{0.837} & \textbf{0.840} & 0.601 \\
    \texttt{BFCL-Base}             & 0.576 & 0.593 & 0.573 & \textbf{0.835} & \textbf{0.834} & 0.606 \\
    ScienceWorld                   & 0.574 & 0.573 & 0.592 & 0.616 & 0.626 & \textbf{0.951} \\
    \bottomrule
  \end{tabular}
\end{table}

\begin{figure}[!ht]
  \centering
  \includegraphics[width=\linewidth]{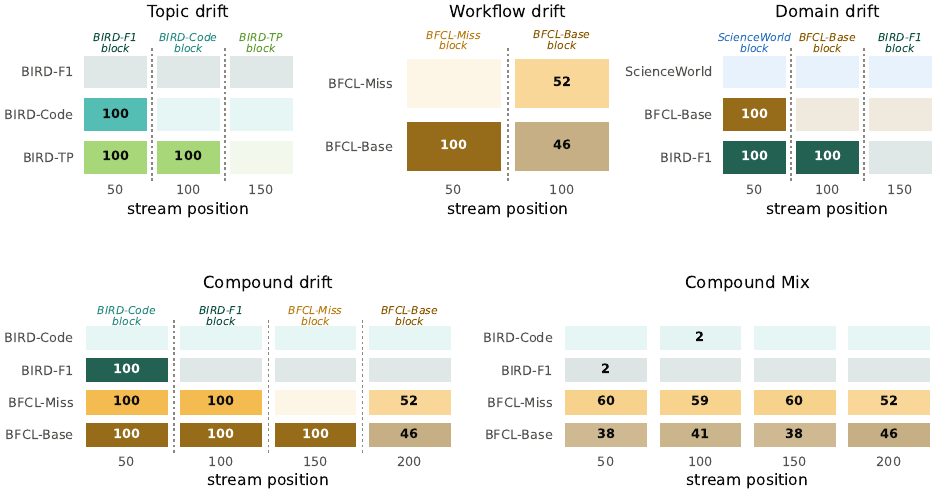}
  \caption{Share (\%) of ReasoningBank evaluation episodes whose injected
  memory entry was written under a task group other than the one
  evaluated, across the four drifting streams and \texttt{Compound Mix}.
  Each row is tinted in the color of the evaluated group, darker with the
  foreign share; empty cells are $0\%$. Dashed lines separate training
  blocks on the drifting streams.}
  \label{fig:rb-retrieval}
\end{figure}

Figure~\ref{fig:rb-retrieval} reports, for every checkpoint and evaluated
task group across the four drifting streams and the interleaved compound
stream (\texttt{Mix}), the share of ReasoningBank evaluation episodes
whose top-$1$ retrieved memory entry was written during a different task
group.

On all four drifting streams, any group evaluated before its own training
block begins ($t \le \sigma_k$) has zero in-group entries in the store and
therefore receives a foreign memory in $100\%$ of evaluation episodes
(the lower-left triangle of $100\%$ cells in
Figure~\ref{fig:rb-retrieval}). Once a group's own block has been
populated, cross-group retrieval drops to $0\%$ whenever the groups have
distinct schemas or domains (topic drift, domain drift, and the BIRD
groups of compound drift). Between \texttt{BFCL-Miss} and
\texttt{BFCL-Base}, however, queries share the same API catalog and
embedding region, causing $52\%$ of \texttt{BFCL-Miss} and $46\%$ of
\texttt{BFCL-Base} episodes to retrieve from the opposite workflow at the
end of workflow drift and compound drift (and $38\%$--$60\%$ throughout
\texttt{Compound Mix}).

\section{Implementation Details}
\label{app:impl}

\begin{table}[!ht]
  \centering
  \small
  \begin{tabular}{ll}
    \toprule
    Parameter & Value \\
    \midrule
    \texttt{temperature} & $1.0$ \\
    \texttt{max\_output\_tokens} & $8192$ \\
    \texttt{thinking\_level} & \texttt{MEDIUM} \\
    \bottomrule
  \end{tabular}
  \caption{Generation parameters.}
  \label{tab:genparams}
\end{table}

\paragraph{Seed harness.} At each step the agent receives the environment
observation and outputs an action in an \texttt{act} block
(Appendix~\ref{app:prompts}). In BIRD, the agent interacts with a
read-only SQLite database via a Bash workspace (with the \texttt{sqlite3}
CLI and per-table CSV descriptions) and finishes by writing its SQL query
to \texttt{/workspace/answer.sql} and calling
\texttt{submit(path="/workspace/answer.sql")}; in BFCL, the block carries
a list of Python function calls executed against stateful multi-turn APIs
(such as a file system, a trading account, and a travel-booking service)
or a \texttt{reply(...)} message to the user; in ScienceWorld, legal text
actions are defined and processed directly by the simulator.

\paragraph{Retrieval.} ReasoningBank embeds its memory items and the
current query with \texttt{gemini-\allowbreak embed\-ding-2}, and retrieves the nearest
stored entry by cosine similarity in that space. It is the only method
with a retrieval step: GEPA and Meta-Harness carry their state in a
shared prompt and workspace respectively, both of which are read in full.

\paragraph{Meta-Harness proposer.} Meta-Harness may rewrite the code of the
harness itself, so its proposer has to read and edit a working codebase. We therefore run the proposer in the opencode
harness~\citep{opencode} rather than in the seed harness, which is the plain
ReAct loop of \S\ref{sec:setup} and offers no such capability. The proposer uses the same Gemini 3.6
Flash~\citep{gemini36flash} as the backend model, with the same generation parameters in Table~\ref{tab:genparams}.

\section{Seed System Prompt and Environment Preambles}
\label{app:prompts}

All three methods share the seed system prompt below and differ in which
of its two anchors they write to. GEPA rewrites the S1 body;
ReasoningBank leaves S1 untouched and injects retrieved memory items at
S2; Meta-Harness may edit both, along with the controller code that
orchestrates them.

\begin{lstlisting}
<S1:REPLACEABLE_SYSTEM_PROMPT>
You are a general agent operating an interactive environment. Solve the task by choosing one executable action at a time from the environment interface described in the user message.

On every turn, respond with exactly one action enclosed in a fenced block whose opening line is exactly ```act. Use this form:

```act
ACTION
```

Only the contents of the first ```act fenced block are sent to the environment. Any content after that block is discarded. Do not invent, predict, or continue an Observation; wait for the environment to return it. A missing, empty, or unclosed ```act block is malformed and still consumes one step. Keep any reasoning concise and place the action itself inside the block.

The episode ends in exactly three cases: the environment declares completion, you execute submit, or the step limit is exhausted.

Follow the environment preamble in the user message for the available action syntax and task rules. Use observations already returned by the environment, verify progress when useful, and submit before the limit when the interface requires submission.
<S2:FIXED_TAIL_INJECTION_SLOT>
\end{lstlisting}

ReasoningBank's items are injected under the header
\texttt{Retrieved Memories}.

\subsection{User Message Template and Environment Preambles}
\label{app:user-prompt}

Every episode opens with one user message assembling the benchmark's
environment preamble, the task statement, and the first observation.
Subsequent turns carry the observation alone, as
\texttt{<OBSERVATION step="$i$">}.

\begin{lstlisting}
<ENVIRONMENT_PREAMBLE>
{environment preamble for this benchmark, reproduced below}
</ENVIRONMENT_PREAMBLE>

<TASK>
{task statement}
</TASK>

<OBSERVATION step="0">
{first observation from the environment}
</OBSERVATION>
\end{lstlisting}

The environment preamble is the only part of the prompt that varies by
benchmark. It states the action interface and the step limit, and no
method may edit it: it is part of the environment, not of the harness.

\paragraph{BIRD.}\mbox{}

\begin{lstlisting}
Environment: isolated SQLite question-answering workspace.

The exact bytes inside each ```act block are interpreted in one of two ways:

1. Any ordinary block is executed as a Bash program in `/workspace`. Standard output, standard error, and the return code are returned as the next observation. Commands time out after 60 seconds. The workspace has no network access.
2. The exact terminal form `submit(path="/workspace/answer.sql")` submits the current contents of that file and ends the episode. No other delivery channel is accepted.

The task database is mounted read-only under `/data/`, and the `sqlite3` command-line shell is installed. Where available, per-table column descriptions are CSV files in the database's `database_description/` directory. Write exactly one SQLite query that answers the question to `/workspace/answer.sql`. Success is determined after submission by strict-prompt-delivery-v1: the submitted query is executed against the database and its result rows are compared with those of a hidden reference query, with a 30-second execution limit.

Step limit: 20 model calls.
\end{lstlisting}

\paragraph{BFCL.}\mbox{} In BFCL the task arrives as the first
\texttt{<USER\_TURN>} block in place of \texttt{<TASK>}, and each later
user turn arrives as a new message once the agent ends the current turn
with \texttt{reply}. The function schemas are filled in per task from the
API classes the task involves.

\begin{lstlisting}
Environment: pinned BFCL v4 local stateful APIs, multi-turn.

The user's requests arrive one turn at a time inside <USER_TURN> blocks.
Observations arrive inside <OBSERVATION> blocks. Each ```act block is one
step and must contain exactly one of:

1. A function-call list in the form [func_a(x=1), func_b(y="v")], one or
   more calls. Calls execute in order against the live APIs; each call's
   result is returned in order in the next observation. Calls cannot be
   nested.
2. The exact form `reply`. It ends the current turn: control returns to
   the user, and their next turn (if any) arrives as the next message.
   `reply` produces no observation. Anything you want to say to the user
   belongs in the text before that act block.

At most 20 steps per turn; at the limit the turn is force-ended. The
episode ends when the final turn ends. There is no submit. Do not assume
the hidden initial backend state beyond what the task and API
observations reveal.

Available function schemas:
<FUNCTION_SCHEMAS>{JSON schemas of the API classes involved in this task}</FUNCTION_SCHEMAS>
\end{lstlisting}

\paragraph{ScienceWorld.}\mbox{}

\begin{lstlisting}
Environment: official ScienceWorld text environment.

An action is the exact text inside the ```act block and is passed to ScienceWorld as one environment action. Available action templates are:

activate OBJ; close OBJ; connect OBJ to OBJ; deactivate OBJ; disconnect OBJ; dunk OBJ in OBJ; eat OBJ; flush OBJ; focus on OBJ; go OBJ; inventory; look around; look at OBJ; look in OBJ; mix OBJ; move OBJ to OBJ; open OBJ; pick up OBJ; pour OBJ in OBJ; put down OBJ; read OBJ; use OBJ on OBJ; wait; wait1.

Use exact object and room names from observations. Open doors or containers before using their contents. Placement uses "move X to Y". `wait` advances ten simulation iterations and `wait1` advances one; physical or biological changes may require them. Many goals require both carrying out the experiment and `focus on X`. The official environment score determines completion. `submit` ends the episode without overriding that score.

Step limit: 100 model calls.
\end{lstlisting}

\end{document}